\documentclass{article}

\usepackage{arxiv}

\usepackage[utf8]{inputenc} % allow utf-8 input
\usepackage[T1]{fontenc}    % use 8-bit T1 fonts
\usepackage{hyperref}       % hyperlinks
\usepackage{url}            % simple URL typesetting
\usepackage{booktabs}       % professional-quality tables
\usepackage{amsfonts}       % blackboard math symbols
\usepackage{nicefrac}       % compact symbols for 1/2, etc.
\usepackage{microtype}      % microtypography
\usepackage{lipsum}
\usepackage{graphicx}
\usepackage{xspace}
\usepackage{xcolor}
\usepackage{amsmath}
\graphicspath{ {./images/} }
\usepackage{ragged2e}
\usepackage{float}
\usepackage{tabularx}
\title{The Saturation Point of Backtranslation in High Quality Low-Resource English–Gujarati Machine Translation}

\author{
 Arwa Arif \\
  Independent Researcher\\
  Dubai, United Arab Emirates\\
  \texttt{arwashafin@gmail.com} \\
}

\begin{document}
\maketitle
\begin{abstract}
Backtranslation (BT) is widely used in low-resource machine translation (MT) to generate additional synthetic training data using monolingual corpora. While this approach has shown strong improvements for many language pairs, its effectiveness in high-quality, low-resource settings remains unclear. In this work, we explore the effectiveness of backtranslation for English–Gujarati translation using the multilingual pretrained MBART50 model. Our baseline system, trained on a high quality parallel corpus of approximately 50,000 sentence pairs, achieves a BLEU score of 43.8 on a validation set. We augment this data with carefully filtered backtranslated examples generated from monolingual Gujarati text. Surprisingly, adding this synthetic data does not improve translation performance and, in some cases, slightly reduces it. We evaluate our models using multiple metrics (BLEU, ChrF++, TER, BLEURT) and analyze possible reasons for this saturation. Our findings suggest that backtranslation may reach a point of diminishing returns in certain low-resource settings and we discuss implications for future research.
\end{abstract}

% keywords can be removed
%\keywords{First keyword \and Second keyword \and More}

\section{Introduction}
Neural Machine Translation (NMT) has seen remarkable progress in recent years, especially with the rise of large multilingual pretrained models like mBART, NLLB and T5. However, many Indian languages including Gujarati remain low-resource, with relatively limited high-quality parallel corpora available for training machine translation systems.

Gujarati, the focus of this study, is an Indo-Aryan language spoken by over 62 million people primarily in the Indian state of Gujarat \textcolor{blue}{(Wikipedia, 2022)} and by communities in the UK, USA and East Africa. It is the sixth most spoken language in India \textcolor{blue}{(Census, 2011)} and the 26th most spoken language in the world \textcolor{blue}{(World Ranking, 2007)}. Despite its wide usage, Gujarati remains a low resource language in the context of natural language processing (NLP). It lacks large scale annotated datasets, robust tools for linguistic preprocessing and high-performing open domain translation models.

Backtranslation (BT) has become one of the most popular techniques for addressing this challenge. It works by translating monolingual sentences from the target language (in our case, Gujarati) back into the source language (English), effectively creating synthetic parallel data. Many studies have shown that this approach can significantly boost translation quality, especially when little supervised data is available.

In this project, we set out to test how much value backtranslation can add when the baseline translation system is already performing well. We use the pretrained \textit{facebook/mbart-large-50-many-to-many-mmt} model and fine-tune it on a clean English–Gujarati parallel corpus of around 50,000 sentences. This baseline achieves a strong BLEU score of 43.8 on a held-out validation set.

We then introduce backtranslated data generated from monolingual Gujarati texts and carefully filter it using token length checks, ratio filters and token similarity heuristics. However, despite the quality of this synthetic data, our experiments show that the model’s performance does not improve. In fact, in some configurations, BLEU drops slightly. This raises an important research question: Is there a saturation point beyond which backtranslation no longer helps, even if the data is high-quality?

Through this work, we aim to answer that question and contribute to a better understanding of where backtranslation is most effective and where it may not be. Our findings are relevant for researchers working on low-resource languages, particularly when using multilingual pretrained models.

\section{Related Works}
\label{sec:headings}
Backtranslation has emerged as a pivotal technique for enhancing neural machine translation (NMT) in low-resource settings, particularly by leveraging monolingual data to generate synthetic parallel corpora. The foundational work of \textcolor{blue}{Sennrich, Haddow and Birch (2016)} demonstrated the effectiveness of backtranslation for NMT, highlighting its utility for exploiting vast monolingual resources when parallel data is scarce\cite{SurveyLRMT}. This approach where target side monolingual data is translated to create synthetic source sentences, has been widely adopted, with \textcolor{blue}{Edunov et al. (2018)} further showing that backtranslation remains beneficial even at scale and that beam search outperforms sampling for generating synthetic data in extremely low-resource scenarios\cite{SurveyLRMT}. \textcolor{blue}{Caswell, Chelba, and Grangier (2019)} introduced a simple yet effective modification by tagging backtranslated data during training, enabling the model to distinguish between natural and synthetic examples and thereby improving overall performance. Variants of backtranslation, such as forward translation \textcolor{blue}{(Zhang and Zong, 2016)} and copying from target to source \textcolor{blue}{(Currey, Miceli Barone, and Heafield, 2017)}, have also been explored, with the latter proving especially useful for named entity translation in low-resource pairs like Turkish–English and Romanian–English\cite{SurveyLRMT}. Iterative backtranslation, as proposed by \textcolor{blue}{Guzmán et al. (2019)} and further investigated by \textcolor{blue}{Hoang et al. (2018), Dandapat and Federmann (2018), Bawden et al. (2019), and Sánchez-Martínez et al. (2020)}, involves successive rounds of translation using intermediate models, but it is noted that diminishing returns are common, with \textcolor{blue}{Chen et al. (2020)} finding that two iterations are often sufficient. Other innovations, such as the use of autoencoder objectives \textcolor{blue}{(Cheng et al., 2016; He et al., 2016; Zhang and Zong, 2016)} and Gumbel softmax for backpropagation through backtranslation \textcolor{blue}{(Niu, Xu, and Carpuat, 2019)}, have been proposed but have shown limited success in low-resource environments. Despite the broad adoption of backtranslation, systematic studies on its saturation and performance limits in low-resource NMT remain scarce, with only a few works such as \textcolor{blue}{Xu et al. (2019) and Chen et al. (2020)} providing empirical evidence of diminishing returns and optimal iteration counts. 

The IIIT-Hyderabad Gujarati-English system (2019) relied on only ~155K parallel sentences, with multilingual training mitigating data scarcity\cite{TheIIIT}. Similarly, English-Marathi experiments achieved marginal gains (+1.2 BLEU) using backtranslation, underscoring the challenge of minimal parallel data\cite{Evaluatingthe}.

Recent studies on specific low-resource language pairs, such as English–Luganda (2025) and English–Manipuri (2022), confirm that while backtranslation can improve translation quality, its effectiveness is highly dependent on the quality and quantity of available data and may plateau or even degrade performance if synthetic data is noisy or redundant\cite{DataAugmentation}\cite{LowResource}. These findings are corroborated by broader surveys which note that, although backtranslation is a cornerstone technique, its practical benefits are context-dependent and may be limited by the inherent challenges of low-resource languages\cite{SurveyLRMT}\cite{ASurvey}.

Studies reveal that low-quality initial translations common in low-resource scenarios, introduce noise, degrading model performance\cite{CombiningSMT}\cite{Evaluatingthe}. For example, mixing statistical MT (SMT) and NMT backtranslated data improved robustness for some languages, but this hybrid approach remains unexplored for Gujarati\cite{CombiningSMT}\cite{Evaluatingthe}.

Prior work establishes backtranslation as beneficial but context dependent. Our findings contribute empirical evidence of its saturation in a high quality baseline system (43.8 BLEU), suggesting that for Gujarati, with its unique morphological and resource constraints alternative strategies like hybrid SMT-NMT data\cite{CombiningSMT} or linguistically informed noise reduction\cite{Investigatingthe} may be necessary to transcend this plateau.

\section{Experimental Setup}
This section outlines the resources and steps involved in preparing data, processing it and training the machine translation models used in this study. Our approach includes a combination of human-translated and backtranslated data, with careful filtering and evaluation to assess the effectiveness of backtranslation in the English–Gujarati setting.

\subsection{Dataset}
We used two main types of data in our experiments: a high-quality human translated parallel corpus and a filtered set of synthetic data generated through backtranslation. The description of all the data used from different sources is given in the Table \ref{tab:dataset} below. The parallel dataset was collected from publicly available sources such as the OPUS project, specifically from sub-corpora like GNOME, Tatoeba and GlobalVoices. After cleaning and removing noisy or duplicate entries, we obtained approximately 50,000 English–Gujarati sentence pairs for training and an additional 10,000 pairs for validation.

To augment this parallel data, we created a backtranslated dataset. We first collected monolingual Gujarati text from open domain sources such as websites, articles and local news. Using our baseline English–Gujarati translation model trained on the parallel corpus, we translated these Gujarati sentences into English to create synthetic English–Gujarati pairs. Initially, we generated about 70,000 such pairs. To ensure quality, we applied a series of filters: a minimum sentence length threshold, a source to target length ratio between 1/3 and 3 and an optional Jaccard similarity check to avoid near-duplicates. After filtering, we retained around 52,000 high-quality synthetic pairs for training.

\begin{table}[ht]
 \caption{Dataset Distribution}
  \centering
  \renewcommand{\arraystretch}{1.5}
  \begin{tabularx}{\textwidth}{l >{\raggedright\arraybackslash}X l >{\raggedright\arraybackslash}X} 
    \toprule
    \textbf{Dataset Type} & \textbf{Resource} & \textbf{Data  Count} & \textbf{Filtering Applied} \\
    \midrule
    Training (Parallel) \newline & OPUS GNOME, Tatoeba (cleaned)     & 50,000         & High quality, deduplicated, filtered \\
    Validation (Parallel) \newline    & OPUS GNOME, Tatoeba (cleaned)   & 10,000         & Held-out, not used in training \\
    Backtranslated (Synthetic)  \newline  & Monolingual Gujarati to Synthetic English    & 52,000   & Filtered for length, ratio and Jaccard \\
    Total (Training + Synthetic) \newline & Combined training data & 102,000 & Used to train backtranslation augmented model \\
    Validation \newline & Combined training data & 10,000 & Used for validation only  \\
    \bottomrule
  \end{tabularx}
  \label{tab:dataset}
\end{table}

\subsection{Preprocessing}
Before training, we applied standard preprocessing techniques to both the parallel and backtranslated datasets. All text was normalized to convert extended Unicode punctuation marks to their standard counterparts. Unprintable characters and unnecessary whitespaces were removed and all sentences were lowercased. Sentences with empty source or target sides were eliminated, as were any examples with obvious malformed or misaligned. We also removed exact duplicates and ensured consistent UTF-8 encoding across the datasets. For the backtranslated data, in particular, we were careful to maintain linguistic quality. We filtered out sentence pairs with significant lexical overlap, extreme length mismatch, or evident hallucination in the English translations. These steps were crucial in ensuring that only clean and meaningful synthetic examples were used for training.

\subsection{Tokenization and Language Setup}
We used the MBart50TokenizerFast tokenizer provided by Hugging Face Transformers, specifically configured for the pretrained \emph{facebook/mbart-large-50-many-to-many-mmt} model. This tokenizer supports over 50 languages using language-specific tokens. For our setup, English inputs were assigned the language code en\_XX and Gujarati targets used gu\_IN. Each sentence was tokenized with a maximum length of 128 tokens. We applied truncation and padding to ensure consistent sequence lengths. The text\_target parameter was used to properly encode the output of the target language for the model. Hugging Face’s \emph{DataCollatorForSeq2Seq} was used to dynamically pad inputs and labels during training, improving efficiency.

\subsubsection{Training Configuration}
We fine-tuned the MBART50 model using Hugging Face’s Seq2SeqTrainer API. Training was carried out for three epochs with a batch size of four for both training and evaluation. The learning rate was set to 1e-5, and a small weight decay of 0.01 was applied to improve generalization. We also used label smoothing with a factor of 0.1 to help the model handle uncertain predictions better. Warmup steps were set to 100 and logging occurred every 50 steps. The model was saved at the end of each epoch to allow for checkpointing and rollback if needed. All experiments were run on a local machine equipped with an NVIDIA RTX-series GPU and a multi-core CPU. The model's output predictions were generated using beam search and we enabled \emph{predict\_with\_generate=True} to ensure accurate translation generation during evaluation.

\subsection{Evaluation}
To evaluate the translation quality of our models, we used multiple metrics beyond BLEU to gain a more comprehensive understanding. BLEU remains the most widely used metric in MT, and our baseline model achieved a strong score of 43.8 on the held-out validation set. However, since BLEU tends to favor exact matches and may not fully reflect the quality of translations for morphologically rich languages like Gujarati, we also evaluated the model using ChrF++, Translation Edit Rate (TER) and BLEURT.

Our model scored 58.3 on ChrF++, indicating strong character-level and morphological overlap between the predictions and the references. The TER score was 25.05, suggesting that the model required relatively few edits to reach the reference translation. To assess semantic adequacy, we used the BLEURT metric, a neural evaluation model pre-trained on human judgments. The BLEURT score for our model was 0.676, supporting the observation that the model maintains strong semantic alignment with the target sentences.

Together, these metrics provide strong evidence that our baseline model already performs at a high level, which may explain why the backtranslated data failed to significantly improve performance.  The results of the different metrics evaluation is given in Table \ref{tab:table}
\begin{table}[H]
 \caption{Evaluation Metrics}
  \centering
  \begin{tabular}{lllll}   
    \cmidrule(r){1-5}
    Model               & BLEU     & ChrF++       & TER        & BLEURT\\
    \midrule
    Baseline (parallel) & 43.8     & 58.3         & 25.1       & 0.676 \\
    + BT (filtered)     & 43.0     & 57.4         & 26.3       & 0.667 \\
    BT only             & 12.0     & 33.2         & 51.6       & 0.412 \\
    \bottomrule
  \end{tabular}
  \label{tab:table}
\end{table}

\section{Methods Used}
This section explains the methodology applied in building and evaluating both the baseline and enhanced models in our English–Gujarati machine translation experiments. We discuss our approach using original parallel data, as well as our use of backtranslated synthetic data for augmentation.

\subsection{Method with Original Parellel Data}
Our initial step was to build a baseline machine translation model using a high quality English–Gujarati parallel corpus. For this, we used the MBART50 multilingual sequence-to-sequence model, a transformer based architecture pretrained on over 50 languages. MBART50 follows a denoising autoencoder approach and supports many-to-many multilingual translation with language specific tokens. It has been shown to perform well even in low resource conditions due to its strong crosslingual representations.

We fine tuned the MBART50 model using approximately 50,000 parallel English–Gujarati sentence pairs. The model was trained using the Hugging Face Transformers library with default encoder-decoder settings and the appropriate language codes (en\_XX for English and gu\_IN for Gujarati). The goal of this baseline system was to establish a reference point for evaluating the effectiveness of additional training data from backtranslation.

Mathematically, the translation process in a neural sequence-to-sequence model can be described as the probability of generating a target sentence trg = ($trg_1$, $trg_2$, ..., $trg_m$) given a source sentence src = ($src_1$, $src_2$, ..., $src_n$) and model parameters $\alpha$. This is expressed as:
\begin{equation}
p(\text{trg} \mid \text{src}; \alpha) = \prod_{k=1}^{m} p\left(\text{trg}_k \mid \text{trg}_{1}, \ldots, \text{trg}_{k-1}, \text{src}; \alpha \right)
\end{equation}
This baseline model achieved a BLEU score of 43.8 on a held-out validation set, which is strong for a low-resource language pair like English–Gujarati.

\subsection{Model with Backtranslated Data}
To address the limitations of the availability of parallel data, we explored backtranslation as a data augmentation technique. Backtranslation allows the use of monolingual data in the target language (Gujarati, in this case) to generate synthetic source–target sentence pairs. The process began by collecting monolingual Gujarati text from web sources and informal content. These sentences were then translated into English using our baseline English–Gujarati model, effectively generating synthetic parallel data. The backtranslated pairs were then filtered to maintain quality. Our filtering criteria included minimum length thresholds, source–target sentence length ratio limits (between 1/3 and 3), and token-level similarity checks. From approximately 70,000 initial synthetic pairs, about 52,000 high-quality pairs remained after filtering.

Then, this synthetic data set was added to the original parallel data to create a new larger training corpus. We experimented with training MBART50 again on this combined dataset to evaluate whether the additional backtranslated data would lead to an improvement in translation quality. However, our results showed that adding synthetic data did not improve performance over the baseline. In fact, BLEU scores either remained the same or slightly declined. This suggests a saturation point, where the model has already learned most of what it can from the available high quality parallel data and the synthetic data, although clean, no longer adds meaningful diversity or depth.

This finding aligns with other recent studies indicating that the effectiveness of backtranslation diminishes when the baseline is already strong or when the backtranslated data closely mirrors existing examples. In our case, English–Gujarati’s strong lexical and syntactic overlap in certain domains may have reduced the marginal gain provided by synthetic examples.

\section{Results}
We evaluated our models using multiple automatic metrics to better understand the impact of backtranslated data on translation quality. Our baseline model, trained only on high-quality parallel data, achieved a BLEU score of 43.8, which is considered strong for a low resource language pair like English–Gujarati. This model also performed well in terms of ChrF++ (58.3) and BLEURT (0.676), indicating a good balance between lexical similarity and semantic adequacy. The TER score of 25.05 further confirmed that the predicted translations required relatively few edits to match the reference sentences.

When we introduced filtered backtranslated data into training, the model's BLEU score dropped slightly to 43.0 and ChrF++ and BLEURT also decreased marginally. Interestingly, the TER increased, suggesting that adding backtranslated data made some translations less direct or more error-prone. The model trained only on synthetic data performed significantly worse, with BLEU falling to 12.0 and other metrics showing similar degradation. This outcome indicates that while the backtranslated data was clean and grammatically sound, it lacked the linguistic diversity or novelty required to further improve the model.

These results clearly demonstrate a performance plateau when adding synthetic data to a well trained baseline system. The model likely had already captured most translation patterns from the clean parallel corpus and the synthetic data despite being filtered, did not introduce new or complementary patterns that could enhance generalization.
The Table \ref{tab:bt-examples} below shows examples of English-Gujarati translation errors, where the 'BT Prediction' fails to match the 'Reference' translation. These errors exemplify the linguistic challenges that continued to persist in the model even after augmentation with backtranslated data.

\begin{table}[ht]
    \caption{Examples of Backtranslation vs. Reference}
    \centering
    \begin{tabularx}{\textwidth}{X X X}
        \toprule
        \textbf{Example (English)} & \textbf{Reference Translation} & \textbf{BT Prediction} \\
        \midrule
        Painting \newline  & \includegraphics[height=1.5em]{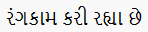} & \includegraphics[height=1.8em]{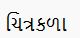} \\
        Show the task preview pane. \newline  & \includegraphics[height=1.5em]{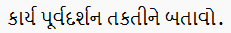} & \includegraphics[height=1.5em]{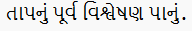} \\
        Please confirm your availability by email. \newline  & \includegraphics[height=2.5em]{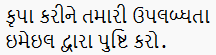} \newline  & \includegraphics[height=2.5em]{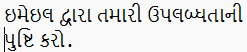} \\
        She felt a strong sense of relief. \newline  & \includegraphics[height=1.5em]{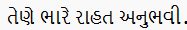} & \includegraphics[height=2.5em]{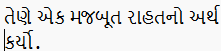} \\
        \bottomrule
    \end{tabularx}
    \label{tab:bt-examples}
\end{table}

\section{Conclusion}
In this study, we explored the use of backtranslation to enhance low resource machine translation for the English–Gujarati language pair using the MBART50 model. While backtranslation is a well established technique for leveraging monolingual data, our experiments showed that it did not improve performance when added to a model already trained on high-quality parallel data. In fact, the addition of filtered synthetic examples led to slightly lower BLEU and ChrF++ scores.
These findings suggest that backtranslation has a saturation point: once a model has been trained on sufficient high quality data, adding synthetic pairs may no longer offer meaningful benefits particularly in domains where the language is repetitive, or the synthetic data closely mirrors existing training examples.
Our results highlight an important consideration for researchers working on low resource languages: backtranslation is not universally beneficial, especially when the baseline system is already strong. Future improvements may require more diverse augmentation techniques, such as paraphrasing, contrastive learning and multilingual knowledge transfer.

\section{Limitations}
One limitation of our study is that we relied on a single pretrained architecture (MBART50) and did not experiment with alternative multilingual or transformer models such as NLLB or IndicTrans2. It's possible that other models may benefit differently from backtranslated data, especially if they are pretrained on a different distribution or designed specifically for Indian languages.
Additionally, we only focused on only one language pair; it would be valuable to test whether similar saturation behavior occurs in other Indian language pairs such as Hindi–Gujarati or Marathi–Gujarati.
In future work, we plan to explore multi-round backtranslation, where models are refined iteratively and human-in-the-loop evaluations to assess how backtranslation affects fluency and adequacy in real usage scenarios. We are also interested in exploring contrastive learning and zero-shot multilingual transfer as alternatives to traditional backtranslation for performance gains.

\bibliographystyle{unsrt}  
\bibliography{references}  %%% Remove comment to use the external .bib file (using bibtex).
%%% and comment out the ``thebibliography'' section.

%%% Comment out this section when you \bibliography{references} is enabled.

\appendix
\section{Appendix}
\subsection{Challenges with Gujarati Language}
The development of high-quality Machine Translation (MT) systems for low resource language pairs like English–Gujarati presents a unique set of challenges that extend beyond simply acquiring parallel data. While Gujarati, an Indic language, has a significant number of speakers, its digital resources for high quality MT training remain relatively sparse compared to major global languages. This inherent data scarcity forms the foundational challenge, as effective neural MT models heavily rely on large, diverse corpora for robust generalization.

\subsection{English-Gujarati Translation Example}
\begin{description}
    \item[Example 1]\leavevmode\par
        \textbf{English:} He built a WiFi door bell, he said.\\
        \textbf{Reference:} \includegraphics[height=1.3em]{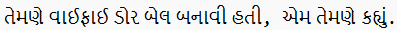}\\
        \textit{(Transliteration: Temne WiFi door bell banavi hati, em temne kahyu.)}\\
        \textbf{Generated by Baseline Model:} \includegraphics[height=1.2em]{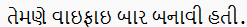}\\
        \textit{(Transliteration: Temne WiFi bari banavi hati.)}\\
        \textbf{Generated by Backtranslation-Augmented Model:} \includegraphics[height=1.4em]{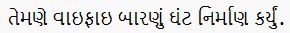}\\
        \textit{(Transliteration: Temne WiFi baranu ghanta nirmaan karyu.)}\\

    \item[Example 2]\leavevmode\par
        \textbf{English:} She is reading a book.\\
        \textbf{Reference:} \includegraphics[height=1.2em]{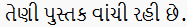}\\
        \textit{(Transliteration: Teṇi pustak vāṇcī rahī che.)}\\
        \textbf{Generated by Baseline Model:} \includegraphics[height=1.2em]{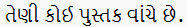}\\
        \textit{(Transliteration: Teṇi koī pustak vāṇce che.)}\\
        \textbf{Generated by Backtranslation-Augmented Model:} \includegraphics[height=1.4em]{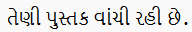}\\
        \textit{(Transliteration: Teṇi pustak vāṇcī rahī che.)}\\

    \item[Example 3]\leavevmode\par
        \textbf{English:} Thank you for your help.\\
        \textbf{Reference:} \includegraphics[height=1.5em]{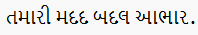}\\
        \textit{(Transliteration: Tamārī madad badal ābhār.)}\\
        \textbf{Generated by Baseline Model:} \includegraphics[height=1.2em]{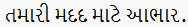}\\
        \textit{(Transliteration: Tamārī madad māṭe ābhār.)}\\
        \textbf{Generated by Backtranslation-Augmented Model:} \includegraphics[height=1.4em]{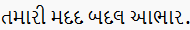}\\
        \textit{(Transliteration: Tamārī madad badal ābhār.)}\\

\end{description}

\subsection{Error Analysis}
A qualitative error analysis of the English–Gujarati translation outputs, particularly from model incorporating back-translated data, revealed several recurring error patterns despite the high baseline BLEU score. As illustrated in the examples presented in Section 5, these errors often indicate the synthetic data's inability to introduce meaningful improvements or capture nuanced linguistic phenomena. Common issues observed included a loss of nuance and specificity, where translations generated more general or less precise terms, failing to capture the full meaning or specific context of the source phrase. We also found instances of lexical and semantic errors, where the model produced fundamentally incorrect word choices or phrases that altered the original meaning. Furthermore, despite a focus on quality, the outputs sometimes suffered from fluency and naturalness issues, exhibiting awkward phrasing or the omission of culturally appropriate polite terms, leading to less natural sounding Gujarati. Finally, a notable challenge was inadequate idiomatic translation, where idiomatic expressions were often translated literally, resulting in ungrammatical output. These persistent error types underscore that while backtranslation can provide data volume, its effectiveness in truly enhancing linguistic diversity and addressing complex translation challenges might be limited once a strong baseline is established with high-quality parallel data.

\end{document}